\documentclass{article}
\usepackage{authblk}
\usepackage[a4paper]{geometry}
\usepackage{graphicx}
\usepackage{amsmath}
\usepackage{amsthm}
\usepackage{fixltx2e}
\usepackage{graphicx}
\usepackage{longtable}
\usepackage{tabu}
\usepackage{amsmath}
\usepackage{textcomp}
\usepackage{marvosym}
\usepackage{wasysym}
\usepackage{amssymb}
\usepackage{verbatim}
\usepackage{mathtools}
\usepackage[normalem]{ulem}
\usepackage{paralist}
\usepackage{tikz}

\theoremstyle{plain}

\newtheorem{theorem}{Theorem}

\theoremstyle{definition}

\newtheorem{definition}{Definition}

\theoremstyle{remark}
\newtheorem{example}{Example}

\newtheorem{remark}{Remark}

\newcommand{\powerset}{\wp}

\newcommand{\rtod}{\tau}

\newcommand{\flaty}{\mathtt{f}}
\newcommand{\preference}{\mathtt{d}}
\newcommand{\reverse}{\mathtt{r}}
\newcommand{\xmode}{\mathtt{x}}

\newcommand{\intro}{\mathcal{I}}
\newcommand{\elim}{\mathcal{E}}

\newcommand{\ABA}{\mathrm{ABA}}
\newcommand{\ABAplus}{\ABA^{+}}
\newcommand{\ABAf}{\ABA^{\flaty}}
\newcommand{\ABAd}{\ABA^{\preference}}
\newcommand{\ABAdwedge}{\ABAd_{\hspace{-.2mm}\wedge} }
\newcommand{\ABAr}{\ABA^{\reverse}}
\newcommand{\ABArwedge}{\ABAr_{\hspace{-.2mm}\wedge} }
\newcommand{\ABAx}{\ABA^{\xmode}}
\newcommand{\ABAxwedge}{\ABAx_{\hspace{-.2mm}\wedge} }
\newcommand{\ABF}{\mathtt{ABF}}

\newcommand{\naive}{\mathsf{naiv}}
\newcommand{\grounded}{\mathsf{grou}}
\newcommand{\preferred}{\mathsf{pref}}
\newcommand{\stable}{\mathsf{stab}}

\newcommand{\semantics}{\mathsf{sem}}

\newcommand{\Val}{\mathbb{V}}
\newcommand{\valued}{\upsilon}

\newcommand{\tolstrut}{%
 \vrule height\dimexpr\fontcharht\font`\A+.1ex\relax width 0pt\relax
}
\DeclareRobustCommand{\textoverline}[1]{%
  \ensuremath{\overline{\mbox{\tolstrut#1}}}%
}

\usepackage{pgfplots}
\newcounter{Gcount}

\def\addlegendimage{\csname pgfplots@addlegendimage\endcsname}
\pgfkeys{/pgfplots/number in legend/.style={%
        /pgfplots/legend image code/.code={%
            \node at (0.125,-0.0225){#1}; 
        },%
    },
}
\pgfplotsset{
every legend to name picture/.style={west}
}

\newcommand{\begprflst}{\begin{itemize}} 
\newcommand{\enprflst}{\end{itemize}}

\makeatletter

\usepackage{tikz}
\usetikzlibrary{shapes,arrows,positioning,backgrounds,fit,calc,shapes.misc}
\title{Assumption-Based Approaches\\ to Reasoning with Priorities \thanks{The research of the authors was supported by a Sofja Kovalevkaja award of the Alexander von Humboldt-Foundation, funded by the German Ministry for Education and Research. A slightly updated version of the paper is forthcoming in the proceedings of AI$^3$.}}
\author{Jesse Heyninck}\author{Pere Pardo} \author{Christian Stra{\ss}er}
\affil{$\{$jesse.heyninck, pere.pardoventura, christian.strasser$\}$@rub.de\\
Institute of Philosophy II, Ruhr Universit\"at Bochum\\
}

\begin{document}
\maketitle

\begin{abstract}
This paper maps out the relation between different approaches for handling preferences in  argumentation with strict rules and defeasible assumptions by offering translations between them. The systems we compare are: non-prioritized defeats, preference-based defeats, and preference-based defeats extended with reverse defeat. We prove that these translations preserve the consequences of the respective systems under different semantics.
\end{abstract}
\section{Introduction}
The aim of this paper is to map out the relation between different approaches for handling preferences in assumption-based argumentation (in short, $\ABA$) \cite{Bondarenko1997}. The orthodox approach in $\ABA$, that we call \emph{direct}, defines defeats (among sets of assumptions) as attacks from assumptions that are at least as preferred as the assumption under attack. The fact that $\ABA$ admits asymmetric contrariness relations, though, makes preference-handling more difficult: this asymmetry is preserved on the level of attacks and then defeats, possibly leading to inconsistencies. In order to re-establish consistency,  the framework $\ABAplus$ was recently proposed in \cite{vcyras2016aba} to handle preferences in $\ABA$. 
$\ABAplus$ adds \emph{reverse defeats} as passive counterparts to \emph{direct defeats}:
if an assumption is attacked from less preferred assumptions a reverse attack is initiated.
Therefore, it seems fruitful to investigate the exact relation between systems that are equipped with a reverse defeat and systems that only make use of direct defeats. In this paper, we contribute to this line of research by studying two questions. First, we investigate under which conditions $\ABA$ equipped with direct but not reverse defeat satisfies the consistency postulate. Thereafter, we investigate the relationship between these two frameworks by providing translations. 

\textbf{Outline of the paper:} In Section \ref{sec:assumpt-based-argum} we review the different versions for $\ABA$ defined by: 
non-prioritized defeats ---i.e. attacks
($\ABAf$), preference-based defeats ($\ABAd$), and {preference-based defeats extended with reverse defeat} ($\ABAr$). 
{In Section \ref{sec:considerations} we motivate the translations by showing first that $\ABAd$ is well-behaved and secondly that $\ABAd$ and $\ABAr$ give rise to incomparable outcomes.}
Then in Section \ref{sec:d2f}, we provide first a translation from $\ABAd$ to $\ABAf$. {In Section \ref{sec:flatintodr} we show $\ABAr$ and $\ABAd$ are conservative extensions of $\ABAf$. This result also extends the translation from Section \ref{sec:d2f} into $\ABAr$.} In Section \ref{sec:r2d}, we complete the cycle by providing a direct translation from $\ABAr$ to $\ABAd$. The contributions of this paper can be summarized in the following diagram:

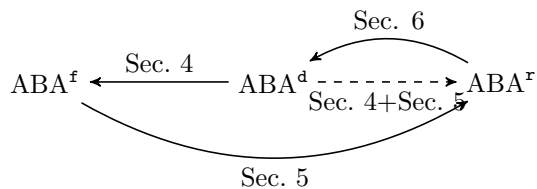
\begin{figure}[h]
\centering
\begin{tikzpicture}
\node (abaf) at (-3, 0) {$\ABAf$};
\node (abad) at (0, 0) {$\ABAd$};
\node (abar) at (3,0) {$\ABAr$};
\path (abar) edge [bend right,->,>=stealth',semithick] node [above] {Sec. \ref{sec:r2d}} (abad);

\path[->,dashed,>=stealth',semithick] 
(abad) edge 
node [below] {{Sec.\ \ref{sec:d2f}+Sec.\ \ref{sec:flatintodr}}} (abar); 
\path[->,>=stealth',semithick] 
(abad) edge node [above] {{Sec.\ \ref{sec:d2f}}} (abaf)
(abaf) edge [bend right=30] node [below] {{Sec.\ \ref{sec:flatintodr}}} (abar);
\end{tikzpicture}
\end{figure}

\section{Assumption-Based Argumentation}
\label{sec:assumpt-based-argum}
ABA, thoroughly described in \cite{Bondarenko1997}, is a formal model on the use of plausible assumptions used ``to extend a given theory'' \cite[p.70]{Bondarenko1997} unless and until there are good arguments for not using (some of) these assumptions. 
 
Inferences are implemented in ABA by means of {rules formulated over a formal language}. Furthermore, defeasible assumptions {are introduced, together with} a contrariness operator to express argumentative attacks. 
We adapt the definition from  \cite{vcyras2016aba} for an $\ABAplus$ assumption-based framework
as follows:
\begin{definition}[Assumption-based framework]\label{deductivesystems}
An \emph{assumption-based framework} is a tuple of the form $\ABF=$ $(\mathcal{L},\mathcal{R}, Ab, {\textoverline{\quad}},\Val, \leq,\valued)$, where:
\begin{itemize}
\item $\mathcal{L}$ is a formal language (consisting of countably many sentences).
\item $\mathcal{R}$ is a set of inference rules of the form $A_1, \ldots, A_n \rightarrow A$ or $\; \rightarrow A$, where $A,A_1\ldots,A_n \in \mathcal{L}$.
\item $Ab \subseteq \mathcal{L}$ is {a non-empty} set of candidate assumptions.
\item $\textoverline{\quad}:Ab\rightarrow \powerset(\mathcal{L})$ is a contrariness operator.
\item The members of $\Val$ are called values and we require that $\Val \neq \emptyset$ and $\Val \cap {\cal L} = \emptyset$.
\item ${\leq}\subseteq \Val\times \Val$ is a  preorder over the values.
\item $\valued:\, Ab\rightarrow \Val$ is a function assigning values to the assumptions\footnote{In \cite{ToniKR}, a preference order ${\leq}\subseteq Ab\times Ab$ is defined directly over the assumptions. It will, however, greatly increase readability to use values to express priorities in this paper. Clearly, these modes of expression are equivalent.}.
\end{itemize}
As usual, we define $\geq$ as the inverse of $\leq$, and define {$\alpha < \beta$ if{f} $\alpha\leq \beta$ and $\beta \not\leq \alpha$.}
{An $\ABF$ without priorities is simply defined as a tuple $\ABF = (\mathcal{L}, \mathcal{R}, Ab, \overline{\phantom{A}})$.}\footnote{If needed, one can identify an $\ABF$ without priorities $(\mathcal{L},\mathcal{R}, Ab, \overline{\phantom{A}})$ with a trivial prioritized $\ABF = (\mathcal{L},\mathcal{R}, Ab, \overline{\phantom{A}}, \Val, \leq, \valued)$ given by $\valued(A) = \valued(B)$ {for all $A,B \in Ab$.}}
\end{definition}

\begin{remark}
{In many publications (e.g.\ 
\cite{Bondarenko1997,Toni2014,ToniKR,ToniNMR}), attention is restricted to so-called \emph{flat} $\ABF$s, i.e.\ $\ABF$s that contain no rules $A_1,\ldots,A_n\rightarrow A$ such that $A\in Ab$. We do not make this assumption but will point to simplifications allowed by it.}
\end{remark}

In some presentations of $\ABA$, deductions are obtained from a set of strict premises $\Gamma \subseteq \mathcal{L}$, a set of plausible assumptions $Ab \subseteq \mathcal{L}$ and a set of rules $\mathcal{R}$. Here we follow \cite{vcyras2016aba}, by 
rewritting each strict premise $A \in \Gamma$ as an empty-bod{ied} rule $\to A$ (contained in the set of rules $\mathcal{R}$).

The previous definition generalizes the contrariness function $\overline{\phantom{A}}: Ab \to \mathcal{L}$ in \cite{vcyras2016aba}, from a single contrary $\overline{A} = B$, to a set of contraries $B_{i} \in \overline{A} = \{B_0, \ldots, B_{k}\}$.
(Although in our examples, for the sake of simplicity, $\overline{A}$ 
will denote an arbitrary member of $\overline{A}$.)
The reason for this {generalization} is to avoid clutter for the translations presented.
\footnote{If one is interested in reducing {a set of contraries $\overline{B}=\{A_1,\ldots,A_n\}$ to a single contrary $\{A_1\}$, one can simply add the rule {$A_i\rightarrow A_1$} for every $1<i\leq n$, cf.\ \cite[p.~109]{Toni2014}.}} 

\begin{definition}[$\mathcal{R}$-deduction]\label{Rdeduction}
Given $\ABF = (\mathcal{L}, \mathcal{R}, Ab, \overline{\phantom{A}}, {\Val, \leq, \valued})$ and a set $\Delta \subseteq Ab$, an {\emph{\(\mathcal{R}\)-deduction}} from \(\Delta\) of \(A\), written \(\Delta \vdash_{\mathcal{R}} A\), is a finite tree where
\begin{enumerate}
\item the root is \(A\),
\item the leaves are either of the form $B$, where $\rightarrow B\in\mathcal{R}$, or elements from \(\Delta\),
\item the children of non-leaf nodes are the conclusions of rules in \(\mathcal{R}\) whose {antecedents} correspond to their {own} parents,
\item \(\Delta\) is the set of all \({B} \in Ab\) that occur as nodes in the tree.
\end{enumerate}
\end{definition}

\begin{remark}
Note that for flat $\ABF$s, if $\Delta\vdash_{\cal R} A$ then $\Delta$ will be the set of all ${B}\in Ab$ occurring as leaves in the tree. The following example shows that for non-flat $\ABF$s we also have to consider non-leaf nodes.
\end{remark}

\begin{example}
\label{ex:1}
Let $\ABF = (\mathcal{L}, \mathcal{R}, Ab, \overline{\phantom{A}}, {\Val, \leq, \valued})$ be given by: \(Ab = \{p,q,r\}\) and the set of rules \(\mathcal{R} = \{ p \rightarrow r, \: p \rightarrow \overline{q}, \: q \rightarrow \overline{r} \}\). Note that there is no deduction \(\{p\} \vdash_{\mathcal{R}} r\) since \(r\) appears as a node in any derivation of \(r\). We have both \(\{r\} \vdash_{\mathcal{R}} r\), whose tree only consists of the root \(r\), and \(\{p,r\} \vdash_{\mathcal{R}} r\) with root \(r\) and unique leaf \(p\). 
\end{example}

Deductions are neither monotonic in the antecedent, e.g. we do not have \(\{p,r,q\} \vdash_{\mathcal{R}} r\) in Ex.~\ref{ex:1}; nor need the antecedent be a closed set of assumptions, e.g.,  \(\{p\} \vdash_{\mathcal{R}} p\) although \(p \rightarrow r \in \mathcal{R}\) in Ex.~\ref{ex:1}.

We define various ways to lift $\leq$ to \emph{sets} of {assumptions}.

\begin{definition}[$\leq$-minimal set]\label{def:minimums}
Given an assumption-based framework $\ABF=( \mathcal{L},\mathcal{R},Ab, {\textoverline{\quad}},\Val,\leq,\valued)$ and $\Delta\subseteq Ab$, we define $\valued(\Delta)=\{\valued(A):~ A\in \Delta\}$ and:
\begin{center}
\begin{tabular}{rl}
 $\min(\Delta)=$ & $\begin{Bmatrix}\alpha\in \valued(\Delta): \nexists \beta\in \valued(\Delta)\mbox{ such that}\ {\beta < \alpha}\end{Bmatrix}$\\[1mm]
 \(\overline{\min}(\Delta)=\) & \(\begin{Bmatrix} \alpha \in \valued(\Delta): \exists \beta \in \min(\Delta) \mbox{ such that\ } \beta\not<\alpha \end{Bmatrix}\). 
\end{tabular}
\end{center}
\end{definition}
The intuition behind $\overline{\min}(\cdot)$ is to \emph{close} $\min$ under incomparable elements: $\overline{\min}(\Delta)$ includes all the elements that are incomparable to at least one element of $\min(\Delta)$.

\begin{definition}[Lifting of $\leq$]\label{lifting}
{Given an assumption-based framework $\ABF=( \mathcal{L},\mathcal{R},Ab, {\textoverline{\quad}},\Val,\leq,\valued)$, $\Delta\cup\{A\}\subseteq Ab$}, we define \footnote{It is not necessary to consider the lifting: \emph{\(\Delta <_{\exists}^{\overline{\min}} A\) if{f} for some \(\valued(B) \in \overline{\min}(\Delta)\), \(\valued(B) < \valued(A)\)}. It can be proved that $<_{\exists}^{\overline{\min}}$ and $<_{\exists}^{\min}$ coincide:  $\Delta <_{\exists}^{\min} A$ iff $\Delta <_{\exists}^{\overline{\min}}A$. {Furthermore, notice that ${<_\forall^{\overline{\min}}}\subseteq {<_\forall^{\min}} \subseteq {<_\exists^{\min}}$.}}
\begin{center}
\begin{tabular}{r@{\qquad}c@{\qquad}r}
\(\Delta <_{\exists}^{\min} A\) & if{f} & for some \({\beta} \in \min(\Delta)\), \({\beta} < \valued(A)\)\\[1.2mm]
\(\Delta <_{\forall}^{\min} A\) & if{f} & for all \({\beta} \in \min(\Delta)\), \({\beta} < \valued(A)\)\\[1mm]
\(\Delta <_{\forall}^{\overline{\min}} A\) & if{f} & for all \({\beta} \in \overline{\min}(\Delta)\), \({\beta} < \valued(A)\)
\end{tabular}
\end{center}
\end{definition}

\begin{remark}\label{remark:liftingscoincidefortotal}
{For any} $\ABF=( \mathcal{L},\mathcal{R},Ab, {\textoverline{\quad}},\Val,\leq,\valued)$ {such that} $\leq$ over $\Val$ is total, {the three liftings} $<_\exists^{\min}$, $<_\forall^{\min}$ and $<_\forall^{\overline{\min}}$ coincide. {From here on, then, when $\leq$ is a total order, we will simply use $<$ to denote any of its liftings: $<^{\min}_\exists, <^{\min}_\forall,<^{\overline{\min}}_\exists$.} The following example shows that all of these lifting principles give rise to different outcomes when considering a non-total preorder.
\end{remark}

\begin{example}\label{ex:2}
Let ${\Val}=\{\alpha_1,\alpha_2,\alpha_3,\alpha_4,\alpha_5\}$ be a set of values with $\valued(A_{i})=  \alpha_i$ and $\leq$ given by the following figure (where a line means that the upper value is more preferred than the lower value, e.g.\ $\alpha_1>\alpha_3$). We have the following:
\begin{figure}
\centering
\begin{tikzpicture}
\node (alpha1) at (0,1) {$\alpha_{1}$};
\node (alpha3) at (.5,0) {$\alpha_{3}$};
\node (alpha5) at (1,1) {$\alpha_{5}$};
\node (alpha4) at (1.5,0) {$\alpha_{4}$};
\node (alpha2) at (2,-1) {$\alpha_{2}$};

\node (text1) at (5.5, 0) {
\begin{tabular}{r@{\,\,}c@{\,\,}l}
$\{A_{2}, A_{3}\}$ & $<_{\exists}^{\min}$ & $A_{4}$\\[1mm]
$\{A_{2}, A_{3}\}$ & $\not<_{\forall}^{\min}$ & $A_{4}$\\[1mm]
$\{A_1,A_3,A_4\}$ & $<_{\forall}^{\min}$ & $A_5$\\[1mm]
$\{A_1,A_3,A_4\}$ & $\not<_{\forall}^{\overline{\min}}$ & $A_5$
\end{tabular}
};

\path[-,semithick]
(alpha1) edge (alpha3)
(alpha5) edge (alpha3)
(alpha5) edge (alpha4)
(alpha4) edge (alpha2);

\end{tikzpicture}
\end{figure}
\end{example}

\begin{definition}[Attack, defeat, reverse defeat]
Given a framework $\ABF=$ $( \mathcal{L},\mathcal{R},Ab, {\textoverline{\quad}},\Val,\leq,\valued)$, a lifting ${<}\in \{<_\exists^{\min},<_\forall^{\min},<_\forall^{\overline{\min}} \}$ and $\Delta \cup \{A\} \subseteq Ab$, 
\begin{center}
\begin{tabular}{lc@{\quad}l}
$\Delta$ \emph{attacks} $A$ (with $\Delta'$) &  if{f} & there is $\Delta'\subseteq \Delta$ such that $\Delta' \vdash_{\mathcal{R}}B$ for some $B\in \textoverline{A}$\\[1mm]
$\Delta$ \emph{$\preference$-$<$-defeats} $A$ & if{f} & $\Delta$ attacks $A$ with some $\Delta'$ such that $\Delta' \not< A$
\end{tabular}
\end{center}
We also say that $\Delta$ \emph{attacks} 
$\Delta'$ if{f} $\Delta$ attacks 
some $A \in\Delta'$; and similarly for $\Delta$ $\preference$-$<$-defeats $\Delta'$. Finally, we say that
\begin{center}
\begin{tabular}{lcl}
$\Delta$ \emph{$\reverse$-$<$-defeats}\footnotemark 
 $\Delta'$ & if{f} & 
$\begin{cases}
\mbox{$\Delta$ $\preference$-$<$-defeats $\Delta'$} &\mbox{ or}\\[1mm]
\begin{matrix}
\mbox{for some $\Delta'' \subseteq \Delta'$ and $A \in \Delta$},\\[1mm]
\hfill\mbox{$\Delta''$ attacks $A$ with $A > \Delta''$}\end{matrix} & \mbox{\emph{(reverse defeat)}} 
\end{cases}$
\end{tabular}
\end{center}

\footnotetext{{We follow \cite{ToniKR} in letting $A$ reverse defeat $\Delta''$ only if $A>\Delta''$. However, we do {not} see any conclusive reason why we should {not} let $A$ reverse defeat $\Delta''$ only if $A\leq \Delta''$. We leave the investigation of this alternative form of $\reverse$-$<$-defeat for future work.}}
\end{definition}  

In the context of ABA without priorities, attack coincides with $\preference$-defeat, so we will sometimes {write} \emph{$\flaty$-defeat} instead of \emph{attack} to avoid confusion. From here on, $\ABAf$, $\ABAd$ and $\ABAr$ denote assumption-based argumentation using, respectively $\flaty$-, $\preference$- and $\reverse$-defeats.

\begin{definition}[${\cal S}$-closure]
Given an $\ABF=( \mathcal{L},\mathcal{R},Ab, {\textoverline{\quad}},\Val,\leq,\valued)$, where $\Delta\subseteq Ab$ and ${\cal S} \subseteq {\cal R}$,
we define:
\vspace{-1mm}
\begin{center}
\begin{tabular}{rc@{\:}l}
{$A\in Cl_{\cal S}(\Delta)$} & if{f} & there is a sequence $A_1,\ldots, A_n$ with $A = A_n$, and for $1\leq i\leq n$\\[1mm]
& & \hfill $A_i\in\Delta$ or $A_i$ {is obtained} by an application of a rule\\[1mm] 
& & \hfill $A_{i_1},\ldots, A_{i_m}\rightarrow A_i$ where $i_1, \ldots, i_m < i$\\[1mm]
$\wp^{\cal S}(\Delta)$ & $=$ & $\{\Delta'\subseteq \Delta:~ \Delta'=Cl_{\cal S}(\Delta')\}$ \hfill (${\cal S}$-closed sets within $\Delta$).
\end{tabular}
\end{center}
Finally, we say that $\Delta$ is ${\cal S}$-\emph{closed}  if{f}  $\Delta\in \wp^{\cal S}(Ab)$.
\end{definition}

The consequences of a given $\ABF$ are determined by the argumentation semantics. On the basis of argumentative attacks, the semantics determine when a set of assumptions $\Delta$ is acceptable. Informally, an acceptable set $\Delta$ should at least not attack itself, and it should be able to defend itself against attacks from other sets of assumptions. Argumentation semantics, originally defined for abstract frameworks in \cite{Dung1995}, have been reformulated for ABA in e.g.\ \cite{Bondarenko1997}. 

\begin{definition}[Argumentation semantics \cite{Bondarenko1997}]\label{def:semantics}
Given a framework $\ABF=( \mathcal{L},\mathcal{R},Ab, {\textoverline{\quad}},\Val,\leq,\valued)$, a lifting ${<}\in \{<_\exists^{\min},<_\forall^{\min},<_\forall^{\overline{\min}} \}$ and sets $\Delta, \Delta' \subseteq Ab$, we define for ${\cal S}\subseteq {\cal R}$ and each $\xmode\in\{\flaty,\preference,\reverse\}$:\vspace{-2mm} 
\begin{flushleft}
\begin{tabular}{l@{\,\,}c@{\quad}l}
$\Delta$ is \emph{$\xmode$-$<$-${\cal S}$-conflict-free} & if{f} & {for no $\Delta'\in \wp^{\cal S}(\Delta)$, $\Delta'$ $\xmode$-$<$-defeats $\Delta$}\\[1mm]
$\Delta$ is \emph{$\xmode$-$<$-${\cal S}$-naive} & if{f} & $\Delta$ is ${\cal R}$-closed and $\subseteq$-maximally $\xmode$-$<$-${\cal S}$-conflict-free\\[1mm]
$\Delta$ \emph{$\xmode$-$<$-${\cal S}$-defends} $\Delta'$ & if{f} & for any $\Delta''\in \wp^{\cal S}(Ab)$ that $\xmode$-$<$-${\cal S}$-defeats $\Delta'$,\\[1mm]
& & \hfill 
there is $\Delta'''\in \wp^{\cal S}(\Delta)$ such that $\Delta'''$ $\xmode$-$<$-defeats $\Delta''$\\[1mm]
$\Delta$ is \emph{$\xmode$-$<$-${\cal S}$-admissible} & if{f} & $\Delta$ is {${\cal R}$-}closed, $\xmode$-$<$-${\cal S}$-conflict-free\\[1mm]
& & \hfill and $\Delta$ $\xmode$-$<$-${\cal S}$-defends every $\Delta'\subseteq \Delta$\\[1mm]
$\Delta$ is \emph{$\xmode$-$<$-${\cal S}$-complete} & if{f} & $\Delta$ is $\xmode$-$<$-${\cal S}$-admissible\\[1mm]
& & \hfill and $\Delta$ contains every $\Delta'$ it $\xmode$-$<$-${\cal S}$-defends\\[1mm] 
$\Delta$ is \emph{$\xmode$-$<$-${\cal S}$-preferred} & if{f}  & $\Delta$ is $\subseteq$-maximally $\xmode$-$<$-${\cal S}$-admissible\\[1mm]
$\Delta$ is \emph{$\xmode$-$<$-${\cal S}$-grounded} & if{f} & $\Delta$ is $\subseteq$-minimally $\xmode$-$<$-${\cal S}$-complete\\[1mm]
$\Delta$ is \emph{$\xmode$-$<$-${\cal S}$-stable} & if{f} & $\Delta$ is {${\cal R}$-}closed, $\xmode$-$<$-${\cal S}$-conflict-free\\[1mm]
& & \hfill and $\Delta$ $\xmode$-$<$-defeats every $A\in Ab\setminus \Delta$
\end{tabular}
\end{flushleft}

{We will denote naive, grounded, preferred resp.\ stable by $\naive$, $\grounded$, $\preferred$, $\stable$.}
For any semantics $\semantics \in \{\naive, \grounded, \preferred, \stable\}$, we define $\xmode\mbox{-}\semantics^{<}_{\cal S}(\ABF)$ as the sets of assumptions that are $\xmode$-$<$-${\cal S}$-$\semantics$, as defined above. \footnote{{Since the order $<$ does not matter in any semantics $\flaty\mbox{-}\semantics^{<}_{{\cal S}}(\ABF)$ or $\flaty$-$<$-${\cal S}$-$\semantics$, 
we will simply write this as $\flaty\mbox{-}\semantics_{{\cal S}}(\ABF)$ and, resp., $\flaty$-${\cal S}$-$\semantics$.}}
\end{definition}

\begin{remark}
In many papers (e.g.\ \cite{Bondarenko1997,vcyras2016aba}), a set $\Delta$ is admissible if it can defend itself from every $\mathcal{R}$-\emph{closed} set of assumptions that defeats $\Delta$.
In the context of priorities, however, this might not always be the most intuitive outcome, as demonstrated by Ex.~\ref{example:closedefeaters}. Therefore, we define both semantics where this requirement is enforced (setting ${\cal S} = {\cal R}$ in Def.~\ref{def:semantics}) and semantics where defeaters are not required to be closed (setting ${\cal S} = \emptyset$ in Def.~\ref{def:semantics}).
\end{remark}

\begin{example}\label{example:closedefeaters}
Let $\ABF = (\mathcal{L},\mathcal{R},Ab,\overline{\phantom{A}}, \Val, \leq,\valued)$ be given by $Ab = \{p,q,r\}$, ${\Val}=\{1,2,3\}$, $\valued(p)=1$, $\valued(q)=2$, $\valued(r)=3$ with $1< 2< 3$ and {$\mathcal{R}=\{ q\rightarrow \overline{p};  r\rightarrow p; r\rightarrow \overline{q} \}$}. For any $\xmode\in\{\preference,\reverse\}$, we have one $\xmode$-$<$-${\cal R}$-complete set: $\{q\}$. To see that $q$ is complete, observe that $\{p,r\}$ is the only closed set that $\xmode$-$<$-defeats $q$. Since $\{q\}$ $\xmode$-$<$-defeats $\{p,r\}$, $\{q\}$ defends itself from $\{p,r\}$. When we move to $\xmode$-$<$-$\emptyset$-complete sets, the situation changes: in that case only $\{p,r\}$ is $\xmode$-$<$-$\emptyset$-complete. To see this, note that $\{r\}$ $\xmode$-$<$-defeats $q$ and $q$ does not $\xmode$-$<$-defeat $\{r\}$. 
\end{example}

One might ask if it is more intuitive to have  $\{p,r\}$ and $\{q\}$ as complete extensions, or just $\{p,r\}$ (which contains the $<$-maximal element $r$). Here we study both options.
This example motivates studying defeaters that are not closed under the full set $\mathcal{R}$,
{as} in Ex.~\ref{example:closedefeaters}. 
In Section \ref{sec:r2d}, we will see an example of an $\ABF$ whose defeaters should 
be
closed under a proper subset of the set of rules $\mathcal{R}$. For another example of semantics parametrized with a set of rules, although for different purposes, see \cite{craven2016argument}.

\section{Some considerations on $\ABAd$ and $\ABAr$}\label{sec:considerations}
In this section we 
motivate the translations given in Sections \ref{sec:d2f} and \ref{sec:r2d}. First we show that $\ABAd$ is well-behaved: 
it satisfies the postulate of \emph{Consistency} under the assumption 
of contraposition. Secondly, we show that even with contraposition, $\ABAd$ and $\ABAr$ might produce incomparable outcomes. 

\subsection{$\ABAd$ and Conflict Preservation}
{In \cite{caminada2007evaluation}}, several \emph{rationality postulates} {were proposed for structured argumentation systems. These postulates describe desirable properties to be satisfied by these systems.} 
The only rationality postulate proposed in \cite{caminada2007evaluation} that is non-trivial for $\ABAd$ and $\ABAr$ {frameworks} is the postulate of \emph{consistency}:
\begin{itemize}
\item no set of assumptions $\Delta$ selected by a given semantics contains an assumption $A$ for which $\overline{A}$ is derivable from $\Delta$
\end{itemize}
(see Theorem \ref{consistencyofabad} {below} for a formal statement). One of the reasons for introducing reverse defeats 
in $\ABAr$ is to avoid {violations of the postulate of consistency by preserving conflicts between assumptions even if the attacking assumptions are strictly less preferred then the attacked assumption. The following example shows that for $\ABAd$, conflicts are not necessarily preserved:}

\begin{example}\label{example:reversedefeat}
Let $Ab=\{p,q\}$, $\mathcal{R}=\{p\rightarrow \overline{q}\}$, $\Val=\{1,2\}$, $v(p)=1$ and $v(q)=2$. Note that $\{p\}$ does not $\preference$-$<$-defeat $q$. As a consequence, $\{p,q\}$ is $\preference$-$<$-${\cal S}$-conflict-free for both $\mathcal{S}=\mathcal{R}$ and $\mathcal{S}=\emptyset$, but at the same time it entails $\overline{q}$.
\end{example}

Accordingly, one might ask under which conditions consistency is preserved in the context of $\ABAd$. 
As in ASPIC$^{+}$ \cite{Prakken2010}, 
one might start by looking at contraposition-like properties. 
\begin{definition}[Contraposition \cite{Toni2014}]
$\ABF=(\mathcal{L},\mathcal{R},Ab,\overline{\phantom{A}},\Val,\leq,\valued)$ is \emph{closed under contraposition} if for every non-empty $\Delta\subseteq Ab$:
\begin{center}
if $\Delta\vdash_{\cal R} C$ for some $C\in \overline{A}$\\[1mm] then for every $B\in\Delta$ it holds that  $\{A\}\cup\Delta\setminus\{B\}\vdash_{\cal R} D$ for some $D\in \overline{B}$.
\end{center}
\end{definition}
{Indeed,} contraposition guarantees consistency, as shown next.
\footnote{In ASPIC$^{+}$ \cite{Prakken2010}, contraposition together with various other conditions are sufficient for consistency. However, for $\ABAd$ it turns out that contraposition alone guarantees consistency.}

\begin{theorem}[Consistency]
\label{consistencyofabad}
Let $\ABF=(\mathcal{L},\mathcal{R},Ab,\overline{\phantom{A}},\Val,\valued,\leq)$ be closed under contraposition. For any ${<}\in \{<_\exists^{\min},<_\forall^{\min},<_\forall^{\overline{\min}} \}$, ${\cal S}\subseteq {\cal R}$ and $\Delta\subseteq Ab$,
\begin{center} 
if $\Delta$ is $\preference$-$<$-${\cal S}$-conflict-free, then for no $A\in Ab$, we have 
$A \in \Delta$ and $\Delta\vdash_{\cal R}\overline{A}$.
\end{center}
\end{theorem}

{The proof of all results in this paper are left out due to space restrictions.}

{Note that Theorem~\ref{consistencyofabad} immediately implies that if $\Delta$ is $\preference$-$<$-${\cal S}$-{naive}, -preferred, -grounded or -stable then for no $A\in Ab$, we have
$A \in \Delta$ and $\Delta\vdash_{\cal R}\overline{A}$.}

\subsection{On the relation between $\ABAd$ and $\ABAr$}
Since {$\ABAr$} reverse-defeat is essentially a form of contrapositive reasoning (cf.\ Section \ref{r2d}), one {might ask} whether a given $\ABF$ closed under contraposition gives the same outcomes under $\ABAd$ {(i.e. without reverse-defeat)} and $\ABAr$. 
A partial answer is given by the following result:
\begin{theorem}\label{corol:d-comp}
Given some $\ABF=(\mathcal{L},\mathcal{R},Ab,\overline{\phantom{A}},\Val,\leq,\valued)$, a set $\mathcal{S}\subseteq \mathcal{R}$ and some ${<}\in \{<_\exists^{\min},<_\forall^{\min},<_\forall^{\overline{\min}} \}$, assume $\ABF$ is closed under contraposition. Then,\vspace{-.5mm}
\begin{itemize}
\item[(1)] any $\preference$-$<$-$\mathcal{S}$-admissible set is an $\reverse$-$<$-$\mathcal{S}$-admissible set; and
\item[(2)] every $\preference$-$<$-$\mathcal{S}$-complete set is a subset of an $\reverse$-$<$-$\mathcal{S}$-complete set. 
\end{itemize}
\end{theorem}

\noindent
However, in general the two approaches produce different outcomes (as can be verified by inspection of \cite[Ex.~13]{Toni2014}): 
\begin{itemize}
\item[(3)] not every $\preference$-$<$-$\mathcal{S}$-complete set is $\reverse$-$<$-$\mathcal{S}$-complete, and moreover 
\item[(4)] not every $\reverse$-$<$-$\mathcal{S}$-admissible set is $\preference$-$<$-$\mathcal{S}$-admissible (or extensible to such a set)
\end{itemize}

\section{Translating $\ABAd$ into $\ABAf$}\label{sec:d2f}
{The} translation from $\ABAd$ into $\ABAf$ essentially embeds the priority ordering $\leq$ over $\Val$ into 
an expanded object language
$\mathcal{L}^{\Val}$. The expanded language $\mathcal{L}^{\Val}$ contains  atoms $A^{\alpha}$ for each atom $A \in \mathcal{L}$ and value $\alpha \in \Val$, and we translate 
\begin{center}
\begin{tabular}{r@{\quad}c@{\quad }c@{\quad}c}
$\tau:$ & $\ABF = (\mathcal{L}, \mathcal{R}, Ab, \overline{\phantom{A}}, \Val, \leq, \valued)$ & $\longmapsto$ &
$\tau(\ABF) = (\mathcal{L}^{\Val}, \tau(\mathcal{R}), \tau(Ab), \overline{\phantom{A}}')$
\end{tabular}
\end{center}
(In fact, we expand the set $\Val$ with a maximum element $\omega$, and abusing notation we denote $\Val \cup \{\omega\}$ again as $\Val$.) With more detail, we translate into {non-prioritized} $\ABA$ frameworks {as follows:}
the assumptions  $A^{\valued(A)} \in \tau(Ab)$ encode the priority $\valued(A)$ of the assumption $A \in Ab$; 
the rules in $\tau(\mathcal{R})$ carry over the antecedents' priorities to the consequent by taking their minimal value; 
and the contrariness operator $\overline{\phantom{A}}'$ (again written as $\overline{\phantom{A}}$ for simplicity)
mirrors the idea of $\preference$-defeat being \emph{an attack that succeeds} by restricting the contrary pairs $A \in \overline{B}$ to those pairs $A^\alpha\in\overline{B^\beta}$ satisfying $\alpha \not< \beta$.

We first discuss the translation for flat, totally ordered frameworks, thereafter explaining the complications when these restrictions are given up.

\subsubsection{Flat Frameworks}

\begin{definition}[Translation $\tau$]\label{d2f-translation}
Where \( \ABF=(\mathcal{L}, \mathcal{R}, Ab, \overline{\phantom{A}},\Val, \leq,\valued)\) is flat and $\Val$ is totally ordered by $\leq$, 
its \emph{translation} $\tau(\ABF) =$ 
\((\mathcal{L}^{\Val}, \tau(\mathcal{R}), \tau(Ab), \overline{\phantom{A}})\) is defined as follows:
\begin{center}
\begin{tabular}{r@{\:}c@{\quad}l}
${\mathcal{L}^{\Val}}$ & $=$ & $\{ A^{\alpha} :~ A \in \mathcal{L}, \alpha \in \Val\}$\\[1mm]
$\tau(A_1,\ldots,A_n\rightarrow A)$ & $=$ & $\begin{Bmatrix} {A_1}^{\alpha_1}, \ldots, {A_n}^{\alpha_n}\rightarrow A^{\alpha} : \alpha \in \min(\alpha_{1},\ldots,\alpha_{n})\end{Bmatrix}$\\[1mm]
$\tau(\rightarrow A)$ & $=$ & $\{ \rightarrow A^{\omega} \}$\\[1mm]

$\tau(\mathcal{R})$ & $=$ & $\bigcup_{r\in\mathcal{R}}\tau(r)$\\[1mm]
$\tau(Ab)$ & $=$ & $\{A^{\valued(A)} :~ A \in Ab\}$\\[1mm]
$A^{\alpha} \in \overline{B^{\valued(B)}}$ & if{f} & $A \in \overline{B}\) and \(\alpha \not< v(B)\)
\end{tabular}
\end{center}
The translation of any set $\Delta \subseteq Ab$ will also be denoted $\tau(\Delta) = \{\tau(A):~ A\in\Delta\}$.
\end{definition}

\begin{example}\label{examplex} Let $\ABF=(\mathcal{L},\mathcal{R}, Ab, \overline{\phantom{A}},\Val,\leq,\valued)$ be given by $Ab=\{p,q\}$, $\mathcal{R}=\{q\rightarrow s\}$ and $s\in\overline{p}$\hspace{.09mm}; and $\Val = \{1,2\}$ with $1 < 2$ and $\valued(p)=1$, $\valued(q)=2$. 

\smallskip
Applying Def.~\ref{d2f-translation} gives us $\tau(\ABF) = (\mathcal{L}^{\Val},\tau(\mathcal{R}), \tau(Ab), \overline{\phantom{A}})$ defined by:
$\tau(Ab)=\{p^1,q^2\}$, $q^2\rightarrow s^{2}\in\tau(\mathcal{R}) $ and $s^{2}\in \overline{p^{1}}$.
\end{example}

\begin{theorem}\hspace{-1.5mm}\footnote{This theorem is a particular case of Theorem~\ref{theorem-fd-complication}, based on a further modification of the translation $\tau$ from Def.~\ref{d2f-translation}, also a particular case of Def.~\ref{d2f-translationBIS} below.}
For any {flat} framework $\ABF = (\mathcal{L},\mathcal{R},Ab,\overline{\phantom{A}}, \Val, \leq, \valued)$ {with a total ordering $\leq$}, any semantics ${\semantics}\in$ $\{\naive,\grounded, \preferred, \stable\}$, any set ${\cal S}\subseteq \mathcal{R}$ and lifting ${<} \in \{<^{\min}_{\exists}, <^{\min}_{\forall}, <^{\overline{min}}_{\forall}\}$,
\begin{center}
\begin{tabular}{r@{\quad}c@{\quad}l}
$\Delta\in {\preference}\mbox{-}{\semantics}^{<}_{{\cal S}}(\ABF)$ & if{f} & 
$\tau(\Delta)\in {\flaty}\mbox{-}{\semantics}_{\tau({\cal S})}(\tau(\ABF))$

\end{tabular}
\end{center}
\end{theorem}

\subsubsection{Non-Flat Frameworks}
The need for modifying Def.~\ref{d2f-translation} in non-flat frameworks is shown next.
\begin{example}\label{ex:d2fnonflat}
Let $\ABF = (\mathcal{L},\mathcal{R}, Ab, \overline{\phantom{A}}, \Val, \leq, \valued)$ be given by: $Ab=\{p,q,r\}$, $\Val =\{1,2,3\}$, $\valued(q)=1$, $\valued(r)=2$, $\valued(p)=3$, with $1 < 2 < 3$,
and $$\mathcal{R}=\begin{Bmatrix}
p\rightarrow q; \quad p,q\rightarrow\overline{r}
\end{Bmatrix}.$$
In $\ABF$ we have that $\{p,q\}$ does not defeat $r$ since $q<r$. Using 
Def.~\ref{d2f-translation}, however, we obtain in $\tau(\ABF)$: $\{p^3\}\vdash_{{\cal R}} q^3$ and thus $\{p^3\}\vdash_{{\cal R}} \overline{r}^3$, so  $\{p^{3},q^{1}\}$ defeats $r^2$. 
{The problem is that the translation from Def.~\ref{d2f-translation} allows us to derive $\overline{r}^3$ from $p^3$ (using $q^3$). This does not mirror the behaviour of $\ABAd$, since there the only deduction of $\overline{r}$ using $p$ would be $\{p,q\}\vdash_{\cal R} \overline{r}$. Since $q$ is used in this deduction, it does not defeat $r$ (since $v(r)>v(q)$). Consequently, the translation from Def.~\ref{d2f-translation} is not adequate for non-flat frameworks.}
\end{example}

\subsubsection{Translation for $\ABAd$ under $<^{\overline{\min}}_\forall$.} 
Let us proceed to define a translation for the lifting $<^{\overline{\min}}_\forall$ (see Def.~\ref{lifting}) which is adequate for frameworks whose preorder $(\Val,\leq)$ is not necessarily total.

\begin{definition}[$\tau$ for $<^{\overline{\min}}_\forall$-$\preference$-defeat]\label{d2f-translationBIS}
Given 
$\ABF = (\mathcal{L}, \mathcal{R}, Ab, \overline{\phantom{A}},\Val, \leq, \valued)$, we define its translation $\tau(\ABF)= (\mathcal{L}^{\Val}, \tau(\mathcal{R}), \tau(Ab), \overline{\phantom{A}})$ as in Def.~\ref{d2f-translation} except for:
\begin{center}
\begin{tabular}{rcl}
$\tau(A_{1},\ldots,A_{n}\rightarrow A)$ & $=$ & $\begin{Bmatrix} A_{1}^{\alpha_{1}}, \ldots, A_{n}^{\alpha_{n}}, A_{1}^{g(A_{1})}, \ldots, A_{n}^{g(A_{n})} \rightarrow A^{\alpha}\end{Bmatrix}_{\alpha_{1},\ldots,\alpha_{n} \in \Val}$\\[1mm]
& & where $\alpha = \min(\{\alpha_{1},\ldots,\alpha_{n}\})$ and\\[1mm]
& & $g(A_{i}) = \valued(A_{i})$ if $A_{i} \in Ab$, and $g(A_{i}) = \alpha_{i}$ otherwise\\[1mm]
$\tau(\mathcal{R})$ & $=$ & $\begin{pmatrix}\bigcup_{r\in \mathcal{R}} \tau(\mathcal{R})\end{pmatrix} \cup 
$$\begin{Bmatrix} A^{\alpha} \rightarrow A^{\valued(A)} : A \in Ab,\, \alpha \in \Val\end{Bmatrix}$ 
\end{tabular}
\end{center}

\end{definition}

Examples like \ref{ex:d2fnonflat} are handled in the translation in Def.~\ref{d2f-translationBIS} by translating rules $r = A_1,\ldots,A_n\rightarrow A$ in a slighlty different way: each antecedent $A_{i}$ in the rule $r$ is translated below into a pair $A_{i}^{\alpha_{i}}, A_{i}^{g(A_{i})}$ of antecedents in $\tau(r)$.

An additional change to Def.~\ref{d2f-translation} can be motivated by Ex.~\ref{ex:d2fnonflat} as well. Indeed, note that the set $\{p^3\}$ would be closed in the translated framework $\tau(\ABF)$ with Def.~\ref{d2f-translation} since $p^3\rightarrow q^3\in \tau(\mathcal{R})$ but ${p^3 \rightarrow q^1} \notin \tau(\mathcal{R})$. However, $\{p\}$ is not closed in $\ABF$ and it can be easily seen that this gives rise to non-adequacies in any of the semantics defined.
This can be fixed by adding new rules in $\tau({\cal R})$ of the form: $A^\alpha\rightarrow A^{\valued(A)}$ for any $A\in Ab$ and $\alpha\in \Val$.

\begin{theorem}\label{theorem-fd-complication}
Let $\ABF = (\mathcal{L},\mathcal{R}, Ab, \overline{\phantom{A}},\Val,\leq,\valued)$ be given, and let $\tau$ be as in Def.~\ref{d2f-translationBIS}. For any semantics ${\semantics}\in \{\naive,\grounded,\preferred,\stable\}$ and any ${\cal S}\subseteq \mathcal{R}$:
\begin{center}
\begin{tabular}{r@{\quad}c@{\quad}l}
$\Delta\in {\preference}\mbox{-}{\semantics}^{<^{\overline{\min}}_{\forall}}_{{\cal S}}(\ABF)$ & if{f} &  $\tau(\Delta)\in {\flaty}\mbox{-}{\semantics}_{\tau({\cal S})}(\tau(\ABF))$
\end{tabular}
\end{center}
\end{theorem}

\subsubsection{Translation for $\ABAd$ under $<^{\min}_\forall$ 
and $<^{\min}_\exists$.} 

For these two liftings, further complications arise, the investigation of which is left for future work.

\section{$\ABAf$ {as a special case of} $\ABAr$ and $\ABAd$}\label{sec:flatintodr}
Suppose {a framework} $\ABF=(\mathcal{L},\mathcal{R}, Ab, \overline{\phantom{A}},\Val,\leq,\valued)$ {is given}, {where $\valued$} is defined by trivial priorities: $\valued(A)=\valued(B)$ for any $A,B\in Ab$. It can be easily verified that both the $\preference$-$<$-$\mathcal{S}$-$\semantics$ and $\reverse$-$<$-$\mathcal{S}$-$\semantics$ sets of assumptions coincide with $\flaty$-$\mathcal{S}$-$\semantics$ sets of assumptions for {all liftings in Def.~\ref{lifting}}. This means that we can capture $\ABAf$ in both $\ABAr$ and $\ABAd$, i.e.\ both $\ABAr$ and $\ABAd$ are conservative extensions of $\ABAf$, generalizing Theorem 5 in \cite{ToniKR}.

\begin{theorem}
Let $\ABF = (\mathcal{L},\mathcal{R}, Ab, \overline{\phantom{A}},\Val,\leq,\valued)$ be given, where  $\valued(A)=\valued(B)$ for any $A,B\in Ab$. 
For any semantics $\semantics\in \{\naive,\grounded,\preferred,\stable\}$, lifting ${<}\in\{<^{\min}_{\forall},<^{\min}_{\exists},{<^{\overline{\min}}_{\forall}}\}$, 
set of rules ${\cal S}\subseteq \mathcal{R}$ and any $\Delta \subseteq Ab$:
\begin{center}
\begin{tabular}{r@{\quad}c@{\quad}l@{\quad}c@{\quad}l}
$\Delta\in {\flaty}\mbox{-}{\semantics}_{{\cal S}}(\ABF)$ 
& if{f} & $\Delta\in {\preference}\mbox{-}{\semantics}^{<}_{{\cal S}}(\ABF)$ & if{f} & {$\Delta\in {\reverse}\mbox{-}{\semantics}^{<}_{{\cal S}}(\ABF)$}.
\end{tabular}
\end{center}
\end{theorem}

\section{Translating $\ABAd$ and $\ABAr$}\label{r2d}\label{sec:r2d}
The translation from $\ABAr$ into $\ABAd$ is based on the idea that reverse-defeat in $\ABAr$ is an instance of \emph{contrapositive} reasoning: whenever 
{$\Delta \vdash_{\mathcal{R}}\overline{A}$} but $\Delta$ is strictly less preferred than $A$, then we should instead reject $\Delta$. This means that {an assumption} can {$\reverse$-defeat} a set of assumptions without attacking {any} particular member of this set; e.g.\ {it can be observed in Ex.~\ref{ex:d2fnonflat} that $r$ $\reverse$-defeats $\{p,q\}$ without $\reverse$-defeating $\{p\}$ or $\{q\}$.} Note that {this mechanism from $\ABAr$ is ruled out in} $\ABAd$: whenever $\Delta$ $\preference$-defeats $\Theta$, {then $\Delta \vdash_{\mathcal{R}} \overline{B}$ for some $B \in \Theta$}. {In order to} capture {it} within $\ABAd$, we proceed {stepwise:} first, we add a conjunction $\wedge$ to $\ABAr$ to make explicit the $\ABAr$ way of defeating a set of assumptions; second, we translate frameworks with conjunction: from $\ABArwedge$ to $\ABAdwedge$. 
These two steps expand the set $\mathcal{R}$ with, first, rules for the introduction and elimination of conjunction and, second, with contrapositive rules.

\begin{figure}[h!]
\centering
\begin{tikzpicture}
\node (abadand) at (0, 0) {$\ABAdwedge$};
\node (abar) at (6,0) {$\ABAr$};
\node (abarand) at (3,0) {$\ABArwedge$};

\path[->,dashed,>=stealth',semithick] 
(abar) edge [bend right] 
node [below] {} 
(abadand); 
\path[->,>=stealth',semithick] 
(abar) edge node [above] {Thm.~\ref{theorem-abf2wedge}}
(abarand)
(abarand) edge  (abar)
(abarand) edge node [above] {Thm.~\ref{theorem-rwedge-dwedge}}
(abadand);
\end{tikzpicture}
\end{figure}
\subsection{The $\ABArwedge$ and $\ABAdwedge$ systems}\label{sec:org1234078}
In the following, let $\Delta'\subseteq _{\mathsf{fin}} \Delta$ denote that $\Delta'$ is a finite subset of $\Delta$, and let $\powerset_{\mathsf{fin}}(\Delta)=\{\Delta':~ \Delta'\subseteq_{\mathsf{fin}}\Delta\}$.
\begin{definition}[Conjunction]\label{def-conjunction}
We say that an 
\(\ABF=(\mathcal{L} ,\mathcal{R},Ab,\overline{\phantom{A}},\Val,\leq,\valued)\) 
{has a \emph{conjunction}} if there is a connective \(\wedge\) such that:
\begin{itemize}
\item[]\(\bigwedge\{A_{1}, \ldots, A_{n}\}\) is in {$Ab$}, for every \(A_{1},\ldots, A_{n} \in {Ab}\) \hfill
($\wedge$-closure of \(Ab, \mathcal{L}\))
\end{itemize}
where \(\bigwedge\{A_1, \ldots, A_n\} = A_1 \wedge \ldots \wedge A_n\),\footnote{In order 
{not to} clutter 
notation we omit brackets and 
{assume the connective} \(\wedge\)
{to be} commutative and associative. 
An enumeration $(A_{0}, A_{1}, \ldots, A_{n},\ldots)$ of the countably-many sentences in $\mathcal{L}$ can be used to define a canonical form for conjunctions, e.g. in increasing order: $\bigwedge\{A_{n_0},\ldots,A_{n_k}\} = A_{n_0} \land \ldots \land A_{n_k}$ for $n_{0} < \ldots < n_{k}$.}
and  for any \(A_{1},\ldots, A_{n} \in {Ab}\) and any $A \in \Delta \subseteq Ab$ with $\bigwedge\Delta \in {Ab}$, the set $\mathcal{R}$ is closed under the following:
\begin{itemize}
\item[] \(A_1, \ldots, A_n \rightarrow \bigwedge\{A_1, \ldots, A_n\}\) 
\hfill ($\wedge$-introduction)\\[.5mm]
\item[] \(\bigwedge \Delta \rightarrow A\) 
\hfill ($\wedge$-elimination) 
\end{itemize}
where $\wedge\intro$ and $\wedge\elim$ denote the sets of $\wedge$-introduction- and $\wedge$-elimination-rules. For any \(\Delta \subseteq Ab\), let 
\begin{center}
\begin{tabular}{rll@{\quad}l}
$\Delta^{\wedge\intro}$ & $= Cl_{\wedge\intro}(\Delta)$ & $= \{ \bigwedge \Delta' :~ {\emptyset\subset  } \Delta' \subseteq_{\sf fin} \Delta\}$ & $where \bigwedge\{A\} = A$\\[1mm]
$\Delta^{\wedge\elim}$ & $= Cl_{\wedge\elim}(\Delta)$ & $= \{ A :~ \bigwedge \Delta' \in \Delta, A \in \Delta' {\cap Ab} \}$ & for $\Delta \subseteq Ab^{\wedge\intro}$
\end{tabular}
\end{center}
\end{definition}
{From now on, we proceed as follows:} if an $\ABF$ has no conjunction we add one, otherwise we use the one present in $\ABF$. In either case, the $\wedge$-closure of the language $\mathcal{L}$ is defined as $\mathcal{L}^{\wedge} = \mathcal{L} \cup Ab^{\wedge\intro}$, and the closure of the set of rules is denoted $\mathcal{R}^{\wedge} = \mathcal{R} \cup \wedge\intro \cup \wedge \elim$. 

\begin{definition}[$\ABF_{\land}$ framework]\label{abf-wedge}
Given an $\ABF=(\mathcal{L} ,\mathcal{R},Ab,\overline{\phantom{A}},\Val,\leq,\valued)$ framework, 
we define {$\ABF_\land=(\mathcal{L}^\land, \mathcal{R}^\land, Ab^{\wedge\intro},\overline{\phantom{A}},\Val^\wedge,\valued^\wedge,\leq^\wedge)$} where \(\mathcal{L}^\land, \mathcal{R}^\land\) and \(Ab^{\wedge\intro}\) are defined as above,\footnote{As stated in Theorem 6, in order to prove equivalence of $\ABAd$ and $\ABAdwedge$ (resp.\ $\ABAr$ and $\ABArwedge$) it is not necessary to define the contrariness operator for conjunctions of assumptions. The situation changes when translating $\ABArwedge$ to $\ABAdwedge$ as is demonstrated (see Def.~13 and Theorem 7) since contraries of conjunctions of assumptions are essential when expressing $\reverse$-defeat in $\ABAdwedge$.}
{$\Val^\wedge=\{V:~ \emptyset\subset V \subseteq_{\mathsf{fin}} \mathbb{V}  \}$}
{$\valued^{\wedge}$ is defined as}: 
{
\begin{itemize}
\item[] $\valued^\wedge(A) = \{\valued(A)\}$\hfill for any $A\in Ab$\\[-2.4mm]
\item[] $\valued^\wedge(\bigwedge\Delta) = \min(\Delta)$ 
\hfill for any $\Delta \subseteq Ab$
\end{itemize}
and $\leq$ is, abusing notation, extended to $Ab^{\wedge\intro}$ as follows:
\begin{itemize}
\item[] $\min(\Delta) = \min(\Delta^{\wedge\elim})$ \: and \:\,  $\overline{\min}(\Delta) = \overline{\min}(\Delta^{\wedge\elim})$ \hfill for any $\Delta \subseteq Ab^{\wedge\intro}$
\end{itemize}
Finally, we extend the lifting of $\leq$ to ${\leq^\wedge}\subseteq\powerset_{\mathsf{fin}}(Ab^{\wedge\intro})\times Ab^{\wedge\intro}$. For $\Theta \subseteq_{\mathsf{fin}} Ab^{\wedge\elim}$,}
\begin{center}
\begin{tabular}{r@{\quad }c@{\quad}l}
$\Delta<^{\min}_\exists \bigwedge\Theta$ & if{f} & for some 
$\beta\in\min(\Delta)$ and 
$\alpha\in\min(\Theta)$
 we have $\beta<\alpha$.\\[1mm]
 $\Delta<^{\overline{\min}}_\exists \bigwedge\Theta$ & if{f} & for some
$\beta\in\overline{\min}(\Delta)$ and every
$\alpha\in\overline{\min}(\Theta)$ 
 we have $\beta<\alpha$.\\[.8mm]
$\Delta<^{\overline{\min}}_\forall \bigwedge\Theta$ & if{f} & for every 
$\beta\in\overline{\min}(\Delta)$ and 
$\alpha\in \overline{\min}(\Theta)$
 we have $\beta<\alpha$.
\end{tabular}
\end{center}
\end{definition}
Where $\xmode\in\{\flaty,\preference,\reverse\}$, we use $\ABAxwedge$ to denote {assumption-based} argumentation for frameworks of type $\ABF_\land$ defined by the notion of $\xmode$-defeat.
For the translation to work, only sets $\Delta$ that are closed under $\wedge\intro$ and $\wedge\elim$ are allowed to $\reverse$-defeat other sets, i.e. $\Delta = Cl_{\wedge\intro \cup \wedge\elim}(\Delta)$. This choice is not arbitrary: the $\wedge$-introduction and -elimination rules are domain independent and {fix} the meaning of the logical connective $\wedge$. 
Indeed, not requiring this would give rise to counter-intuitive examples,
like being able to \emph{argue against $p\land q$} but unable to \emph{defend against $\{p,q\}$}.
\begin{example}\label{ex-abf-wedge}
Let $\ABF = (\mathcal{L}, \mathcal{R}, Ab, \overline{\phantom{A}}, \Val, \leq, \valued)$ be given by: $Ab=\{p,q,r\}$; $\mathcal{R}=\{p,q\rightarrow s\}$ with $\{s\}=\overline{r}$ and $\Val = \{1,2,3\}$ with $1<2<3$ and $\valued(p) = 1$, $\valued(q) = 2$, $\valued(r) = 3$.
Applying Def.~\ref{abf-wedge}, we obtain the following $\ABF_\wedge$ framework:\vspace{-2mm} 
\footnotesize
\begin{center}
\begin{tabular}{c@{\hspace{1mm}}c@{\hspace{1.2mm}}c}
$\mathcal{L}^\wedge= \mathcal{L} \cup Ab^{\wedge\intro}$ & $\mathcal{R}^\wedge=\mathcal{R}\cup \wedge\intro \cup \wedge\elim$ & $\valued^{\wedge}$\\[1mm]
\hline
\\[-2mm]
$\begin{matrix}
\begin{Bmatrix}
\begin{matrix}
s, \quad \ldots\\[1mm]
p, \quad q, \quad r,\\[1mm]
p \land q, \quad p \land r,\\[1mm]
q \land r, \quad p \land q \land r
\end{matrix}
\end{Bmatrix}
\end{matrix}$
& 
$\begin{Bmatrix}
\begin{matrix}
p, q \to s\quad 
p,q \to p \land q\quad
p, r \to p \land r\\[1mm]
q, r\to q \land r\quad
p,q,r \to p\land q \land r\\[1mm]

p \land q \to p\quad
p \land q \to q\quad
p \land r \to p\\[1mm]
p \land r \to r\quad
q \land r \to q\quad
q \land r \to r
\end{matrix}
\end{Bmatrix}$
&
$\begin{matrix}
\valued^{\wedge}(p \land q) ={\{ 1\}}\\[1mm]
\valued^{\wedge}(p \land r) = {\{1\}}\\[1mm]
\valued^{\wedge}(p \land q \land r) = {\{1\}}\\[1mm]
\valued^{\wedge}(q \land r) = {\{2\}}\\[1mm]
\end{matrix}$
\end{tabular}
\end{center}

\end{example}

\begin{theorem}\label{theorem-abf2wedge}
For any $\ABF = (\mathcal{L}, \mathcal{R}, Ab, \overline{\phantom{A}}, \Val, \leq, \valued)$, $\semantics\in \{\naive, \grounded, \preferred, \stable\}$, a defeat type $\xmode\in \{\preference,\reverse\}$, lifting ${<}\in \{<_\exists^{\min},<_\forall^{\min},<_\forall^{\overline{\min}} \}$ and {any set $\mathcal{S}$ with $(\wedge\intro\cup\wedge\elim) \subseteq \mathcal{S} \subseteq \mathcal{R}^{\wedge}$}, we have:
\begin{center}
\begin{tabular}{r@{\quad}c@{\quad}l}
$\Delta\in {\xmode}\mbox{-}{\semantics}^{<}_{{\cal S}}(\ABF)$ & if{f} & $\Delta^{\wedge\intro}\in {\xmode}\mbox{-}{\semantics}^{<}_{{\cal S}}(\ABF_\wedge)$
\end{tabular}
\end{center}
\end{theorem}

\subsection{Translating $\ABArwedge$ to $\ABAdwedge$}
\label{sec:org69420cc}

When translating $\ABAr$ to $\ABAd$ we first translate $\ABAr$ to $\ABArwedge$ and subsequently to $\ABAdwedge$. In the previous section we have shown how to do the former step, now we show how to translate $\ABArwedge$ to $\ABAdwedge$.

For this, we extend the language with new formulas $A^{\lnot}$ for any $A\in Ab^\wedge$ that will function as an additional contrary of the assumption $A$ (in Ex.~\ref{ex:extendedlanguage} we will motivate this extension). Furthermore we add contrapositive rules. In particular, whenever:
\begin{itemize}
\item $B\in\overline{C}$ can be derived from $A_1,\ldots,A_n$, and
\item $C$ is strictly preferred (for some ${<}\in \{<_\exists^{\min},<_\forall^{\min},<_\forall^{\overline{\min}} \}$) over $\{A_1,\ldots,A_n\}$, 
\end{itemize} 
we add the rule $C\rightarrow D$, where {$D = (A_{1}\land \ldots \land A_{n})^{\neg}$} is the contrary of $A_1\land \ldots \land A_n$. 

\begin{definition}[$\ABF_{\wedge\neg}$ framework]\label{abf-wedge-neg}
Given $\ABF_{\wedge} = (\mathcal{L}^{\wedge},\mathcal{R}^{\wedge}, Ab^{\wedge},\overline{\phantom{A}},\Val,\leq,\valued)$, we define its translation as $\ABF_{\wedge\neg}=(\mathcal{L}^{\wedge\neg},\mathcal{R}^{\wedge\neg}, Ab^\wedge,\widetilde{\phantom{a}},{\Val^\wedge,\valued^\wedge,\leq^\wedge})$, defined as follows:
\begin{center}
\begin{tabular}{rcl}
$\mathcal{L}^{\wedge\neg}$ & $=$ & $\mathcal{L}^{\wedge} \cup \{ A^{\neg} :~ A \in Ab^{\wedge}\}$\\[1mm]
$\mathcal{R}^{\wedge\neg}$ & $=$ & $\mathcal{R}^{\wedge} \cup 
\begin{Bmatrix} C \to (A_{1}\wedge\ldots\wedge A_{n})^{\neg} \, :\, \begin{array}{r}
 \{A_{1},\ldots,A_{n}\} \vdash_{\mathcal{R}^{\wedge}} B \mbox{ and}\\[1mm] B \in \overline{C}\mbox{ and}\\[1mm] \{A_{1},\ldots,A_{n}\} < C\end{array}  \end{Bmatrix}$\\[1.5mm]
$\widetilde{A}$ & $=$ & $\overline{A} \cup \{A^{\neg}\}$, \hfill for any {$A \in Ab^{\wedge}$}
\end{tabular}
\end{center}
\end{definition}


\begin{example}[Cont'd]\label{}
Let $\ABF_{\wedge}$ be as in Ex.~\ref{ex-abf-wedge}.
Note that $\{r\}$ $\reverse$-defeats $\{p,q\}$. Applying Def.~\ref{abf-wedge-neg} to $\ABF_{\wedge}$, we obtain the translation $\ABF_{\wedge\neg}$ given by: 
\begin{flushleft}
\begin{tabular}{rcl}
$\mathcal{L}^{\wedge\lnot}$ & $=$ & $\mathcal{L}^{\wedge} \cup
\begin{Bmatrix}
\begin{array}{@{\:}l@{\quad}l@{\quad}l@{\quad}l@{\quad}l@{\quad}l@{\quad}l@{\quad}l@{\:}}
p^{\neg},& q^{\neg}, & r^{\neg},& (p\land q)^{\neg}, & (p\land r)^{\neg}, & (q\land r)^{\neg}, & (p\land q\land r)^{\neg}
\end{array}
\end{Bmatrix}$
\end{tabular}\\[1mm]
\begin{tabular}{rcl}
$\mathcal{R}^{\wedge\lnot}$ & $=$ & $\mathcal{R}^\land\cup \{r\rightarrow (p\land q)^{\neg}\}$.
\end{tabular}\\[.7mm]
{
\hspace{1.3mm}$\widetilde{p} = \overline{p} \cup \{p^{\neg}\}$\quad
$\widetilde{q} = \overline{q} \cup \{q^{\neg}\}$\quad
$\widetilde{r} = \{s, r^{\neg}\}$\quad
$\widetilde{p \land q} = \{(p\land q)^{\neg}\}$\quad
\ldots
}
\end{flushleft}
\end{example}

The following example shows why we cannot simply add $C\rightarrow \overline{A}$, where $A\vdash_{\cal R^\wedge}\overline{C}$ and $A<C$, as a contrapositive rule:
\begin{example}\label{ex:extendedlanguage}
Let $\ABF=(\mathcal{L}, \mathcal{R}, Ab, \overline{\phantom{A}}, \Val, \leq, \valued)$ be given by: $Ab=\{p,q\}$; $\mathcal{R}=\{ p\rightarrow \overline{q}; \: \overline{p}\rightarrow r\}$ and $\valued(p) < \valued(q)$. Suppose now that we would add $q\rightarrow \overline{p}$ instead of $q\rightarrow p^\lnot$ to $\mathcal{R}^{\wedge\lnot}$. In that case we would have the deduction $\{q\}\vdash_{\mathcal{R}^{\wedge\lnot}} r$. Since $\{q\}\not\vdash_{\mathcal{R}} r$, this would render the translation inadequate. 
\end{example}

\begin{theorem}\label{theorem-rwedge-dwedge}
For any $\ABF_{\wedge} = (\mathcal{L}^{\wedge},\mathcal{R}^{\wedge}, Ab^{\wedge},\overline{\phantom{A}},{\Val^\wedge,\valued^\wedge,\leq^\wedge})$, semantics ${\semantics}\in \{\naive, \grounded, \preferred, \stable\}$, lifting ${<}\in \{<_\exists^{\min},<_\forall^{\min},<_\forall^{\overline{\min}} \}$ and any set ${\cal S}$ with $(\wedge\intro\cup\wedge\elim)\subseteq \mathcal{S}\subseteq {\cal R}^\wedge$:
\begin{center}
\begin{tabular}{r@{\quad}c@{\quad}l}
$\Delta\in {\reverse}\mbox{-}{\semantics}^{<}_{{\cal S}}(\ABF_\wedge)$ & if{f} & $\Delta\in {\preference}\mbox{-}{\semantics}^{<}_{{\rtod}(\cal S)}(\tau(\ABF_{\wedge}))$.
\end{tabular}
\end{center}
\end{theorem}

The translation proposed here makes use of the (meta-)notion of a deduction, i.e.\ $C \to (A_{1}\wedge\ldots\wedge A_{n})^{\neg}\in\mathcal{R}^{\wedge\lnot}$ iff  $\{A_{1},\ldots,A_{n}\} \vdash_{\mathcal{R}^{\wedge}} B$ and $B>\{A_{1},\ldots,A_{n}\} $. It would perhaps be more elegant to have contraposition on the level of the rules rather than to base it on the derivability relation $\vdash_{\mathcal{R}}$.
Such a proposal, however, runs into additional complications. The following example demonstrates why we cannot just replace $\mathcal{R}^{\wedge\lnot}\setminus\mathcal{R}^\wedge$ by $\{A\rightarrow (\bigwedge_{i=1}^n A_i)^\lnot :~ A_1,\ldots,A_n\rightarrow \overline{A}\in\mathcal{R}\}$:
\begin{example}
Let $Ab=\{p,q\}$ and $\mathcal{R}=\{p\rightarrow s,s\rightarrow \overline{q}\}$. Note that we can't add $q\rightarrow s^\lnot$ to $\mathcal{R}$ since $s^\lnot$ is not defined (since $s\not\in Ab$). Of course one could extend the language with $A^\lnot$ for $A\in\mathcal{L}\setminus Ab$. However, we leave the investigation of this proposal for future research.
\end{example}

\section{Related Work}
In \cite{Bondarenko1997,Toni2008b} ways of expressing priorities in the object language of ABA were proposed. In our contribution we demonstrated how this idea can be utilised to express the ways priorities are handled in $\ABAd$ and $\ABAr$ in the basic (non-prioritized) $\ABA$ framework of \cite{Bondarenko1997}. In \cite{ToniNMR} it was shown that (a special case of) $\ABAr$ conservatively extends $\ABA$ from \cite{Bondarenko1997}. We have generalized this result to $\ABAd$ and by translating both to $\ABA$ we have shown that the expressive power of the three frameworks (w.r.t.\ the standard semantics) is the same.

\subsubsection{On the relation between $\ABAplus$ and $\ABAr$.}
The idea of reverse-defeat was first introduced in \cite{ToniKR} in the context of $\ABAplus$. In this subsection we will discuss the various versions of $\ABAplus$ and their relation to $\ABAr$ as defined in this paper.

In \cite{ToniKR} we find the following definition of defence and a corresponding notion of admissibility (for flat assumption-based frameworks):
\begin{definition}[Defence, admissibility in $\ABAplus$]
Define, for $\Delta \cup \{A\} \subseteq Ab$,\vspace{-.5mm}
\begin{center}
\begin{tabular}{r@{\hspace{3mm}}c@{\hspace{3mm}}l}
$\Delta$ \emph{defends$^+$} $A$ 
&  if{f} & $\Delta$ $\reverse$-$<^{\min}_\exists$-defeats every $\Theta\subseteq Ab$ that $\reverse$-$<^{\min}_\exists$-defeats $\{A\}$.\\[1mm] 
$\Delta$ is \emph{admissible$^+$} & iff & 
$\Delta$ is $\reverse$-$<^{\min}_\exists$-conflict-free  and defends$^+$ every $A\in \Delta$.
\end{tabular}
\end{center}
\end{definition}

This definition gives counter-intuitive outcomes as {shown next}. 
\begin{example} Let $\ABF$ be given by
$\mathcal{R}=\{p,q\rightarrow \overline{r}\}$, $Ab=\{p,q,r\}$ and $\valued(p) < \valued(q) < \valued(r)$. Note that $\{r\}$ $\reverse$-$<^{\min}_\exists$-defeats $\{p,q\}$ but it defeats neither $\{p\}$ nor $\{q\}$. Consequently, $\{p,q\}$ is admissible, even though $\{r\}$ defeats $\{p,q\}$ without any defence from $\{p,q\}$.
\end{example}

This definition has been changed in an online manuscript \cite{vcyras2016aba}, where we find definitions for admissible and complete sets for which it can be routinely checked {to be} equivalent to the definition of a $\reverse$-$<^{\min}_{\exists}$-${\cal R}$-complete set. Thus, the results in this paper cover the definitions of \cite{vcyras2016aba} as a special case.

\subsubsection{On the relation between $\ABAd$ and ASPIC$^+$.}
In \cite{Prakken2010}, it was proven that flat $\ABF$s can be straightforwardly translated into ASPIC$^+$. However, when prioritized, non-flat $\ABF$s come into play, this is not the case any more. In {our} Ex.~\ref{example:closedefeaters}, it can be verified that ASPIC$^+$ would give rise to the same outcome as the $\preference$-$<$-$\emptyset$-preferred semantics (which has not yet been considered in the literature). For flat $\ABF$s, our results together with those of \cite{Prakken2010} show that $\ABAd$ and $\ABAr$ can be expressed in ASPIC$^+$. Further investigations {into the relation} between (non-flat) $\ABA$, $\ABAd$ and $\ABAr$ on the one hand and ASPIC$^+$ on the other hand remain open for future research.

\section{Conclusion}

This paper contains two main results. First, we showed that $\ABAd$, a system that was not investigated until now, satisfies the consistency postulate when $\ABF$s are closed under contraposition. This result was to be expected in view of analogous results for ASPIC$^+$. Second, we investigated translations between $\ABAf$, $\ABAd$ and $\ABAr$. We showed that $\ABAd$ can be translated into $\ABAf$. 

{Intuitively, this means that --at least for $\ABA$-- structured argumentation is as expressive as \emph{prioritized} structured argumentation (the former encoding on the object level the priorities that the latter system expresses on the meta-level).} This result {is along the same line of} results in abstract argumentation, where it {has been} shown that enhancements of Dung's original frameworks are reducible to the original framework by adding extra arguments and attacks \cite{boella2009meta}. This does not mean that the enhancements are not useful {e.g. for ease of representation or succintness}. {These kind of results offer} meta-theoretic insights into the foundations of non-monotonic reasoning, {while at the same time showing } that automatic reasoning systems devised for the base systems can still be used (under translation) for the enhancements.

More specifically, {one can summarize} the insights exposed by our technical results, {as well as point out} possibilities for future work {arising from them}: 

\smallskip
\noindent
(1) We have shown that the way priorities are handled in $\ABAd$ can be expressed in the object language. We plan to investigate whether this is also the case for e.g.\ ASPIC$^+$, where priorities are handled in a similar way to $\ABAd$. Furthermore, expressing priorities in the object language will facilitate research on reasoning \emph{about} priorities in $\ABA$.

\smallskip
\noindent
(2) Our results show that reverse attacks can be expressed by means of contraposition in $\ABAd$ given that the language is logically sufficiently expressive (i.e. contains a conjunction). Consequently, our results clarify the status of reverse attacks w.r.t. more orthodox approaches to handling priorities in structured argumentation. 

\smallskip
\noindent
(3) {While studying} the relation between $\ABAd$ and $\ABAr$, we have shown that adding a conjunction to these frameworks does not change the consequences of a given $\ABF$. In future work, we would like to investigate the effect of closing $\ABF$s under other logical connectives such as disjunction, implication or different forms of negation. Similar research has been done for adaptive logics \cite{van2014adaptive}. 

\smallskip
\noindent
(4) We would also like to point out that in this paper we have made several generalizations w.r.t. $\ABA$ as it is found in the literature. For example, we ``parametrized'' {the} semantics in Def.~\ref{def:semantics} {with \emph{sets of rules}} ${\cal S}$ and {\emph{liftings}} $<$ (of which only one, $<^{\min}_\exists$, {had been} investigated in the literature for $\reverse$-defeats). We think that these generalizations of existing $\ABA$ semantics offer further avenues for {research}, {e.g.} properties for non-monotonic reasoning (cf.\ \cite{ToniNMR}) and rationality postulates.

\bibliographystyle{plain}
\bibliography{KR-final.bib}

\end{document}